\documentclass[10pt,twocolumn,letterpaper]{article}

\usepackage[accsupp]{axessibility}
\usepackage{iccv}
\usepackage{times}
\usepackage{epsfig}
\usepackage{graphicx}
\usepackage{amsmath}
\usepackage{ntheorem}
\usepackage{amssymb}
\usepackage{enumitem}
\usepackage{xcolor}
\usepackage{adjustbox}
\usepackage{multirow}
\usepackage{booktabs}
\usepackage{diagbox}
\usepackage{placeins}

\makeatletter
\newif\if@restonecol
\makeatother

\usepackage[ruled,vlined]{algorithm2e}
\usepackage{algpseudocode}

\newtheorem{theorem}{Theorem}[section]

\newtheorem{lemma}[theorem]{Lemma}

\newtheorem{example}[theorem]{Example}

\usepackage[breaklinks=true,bookmarks=false]{hyperref}
\hypersetup{colorlinks, breaklinks, citecolor=[rgb]{0, 0, 0},
anchorcolor=[rgb]{0,0,0}, linkcolor=[rgb]{0,0,0},
}
\iccvfinalcopy 

\def\iccvPaperID{2916} 

\ificcvfinal\fi

\begin{document}
\newcommand{\mix}{MixMix}

\title{MixMix: All You Need for Data-Free Compression Are Feature and Data Mixing}

\author{Yuhang Li$^{*\dag}\ \ \ \ \ \ $ Feng Zhu$^{\dag}\ \ \ \ \ \ $ Ruihao Gong$^{\dag}\ \ \ \ \ \ $ Mingzhu Shen$^{\dag}\ \ \ \ \ \ $ Xin Dong$^{\ddag}\ \ \ \ \ \ $\\
Fengwei Yu$^{\dag}\ \ \ \ \ \ $ Shaoqing Lu$^{\dag}\ \ \ \ \ \ $ Shi Gu$^*$ \\
{$^*$UESTC $\ \ \ \ \ \ \ $ $^{\dag}$SenseTime Research$\ \ \ \ \ \ \ $ $^{\ddag}$Harvard University}\\
}

\maketitle
\ificcvfinal\thispagestyle{empty}\fi

\begin{abstract}

User data confidentiality protection is becoming a rising challenge in the present deep learning research. Without access to data, conventional data-driven model compression faces a higher risk of performance degradation. Recently, some works propose to generate images from a specific pretrained model to serve as training data. 
However, the inversion process only utilizes biased feature statistics stored in one model and is from low-dimension to high-dimension. As a consequence, it inevitably encounters the difficulties of generalizability and inexact inversion, which leads to unsatisfactory performance. 
To address these problems, we propose MixMix based on two simple yet effective techniques: (1) \textbf{Feature Mixing}: utilizes various models to construct a universal feature space for generalized inversion; (2) \textbf{Data Mixing}: mixes the synthesized images and labels to generate exact label information. We prove the effectiveness of MixMix from both theoretical and empirical perspectives. Extensive experiments show that MixMix outperforms existing methods on the mainstream compression tasks, including quantization, knowledge distillation and pruning. Specifically, MixMix achieves up to 4\% and 20\% accuracy uplift on quantization and pruning, respectively, compared to existing data-free compression work.

\end{abstract}


\section{Introduction}
To enable powerful deep learning models on the embedded and mobile devices without sacrificing task performance, various model compression techniques have been discovered. For example, neural network quantization~\cite{gong2019dsq, hubara2017qnn, li2019apot, zhu2016ttq} converts 32-bit floating-point models into low-bit fixed point models and benefits from the acceleration of fixed-point computation and less memory consumption. Network pruning~\cite{dong2017obslayer, han2015deepcompress, theis2018fisherprune} focuses on reducing the redundant neural connections and finds a sparse network. Knowledge Distillation (KD)~\cite{hinton2015distilling, polino2018distillation} transfers the knowledge in the large teacher network to small student networks. 

\begin{figure}[t]
   \centering
   \includegraphics[width=\linewidth]{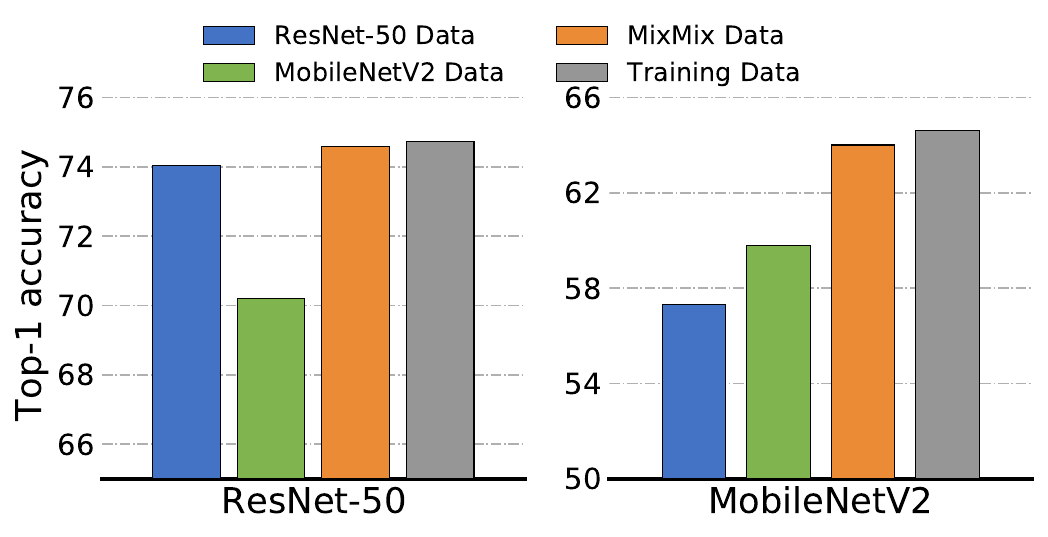}
   \caption{An overview of the results of post-training data-free quantization. Each color bar denotes a data source. Data inverted from ResNet-50 encounters an accuracy deficit on MobileNetV2.}
   \label{fig_cross_validation}
\end{figure}
\begin{figure*}[t]
    \centering
    \includegraphics[width=\linewidth]{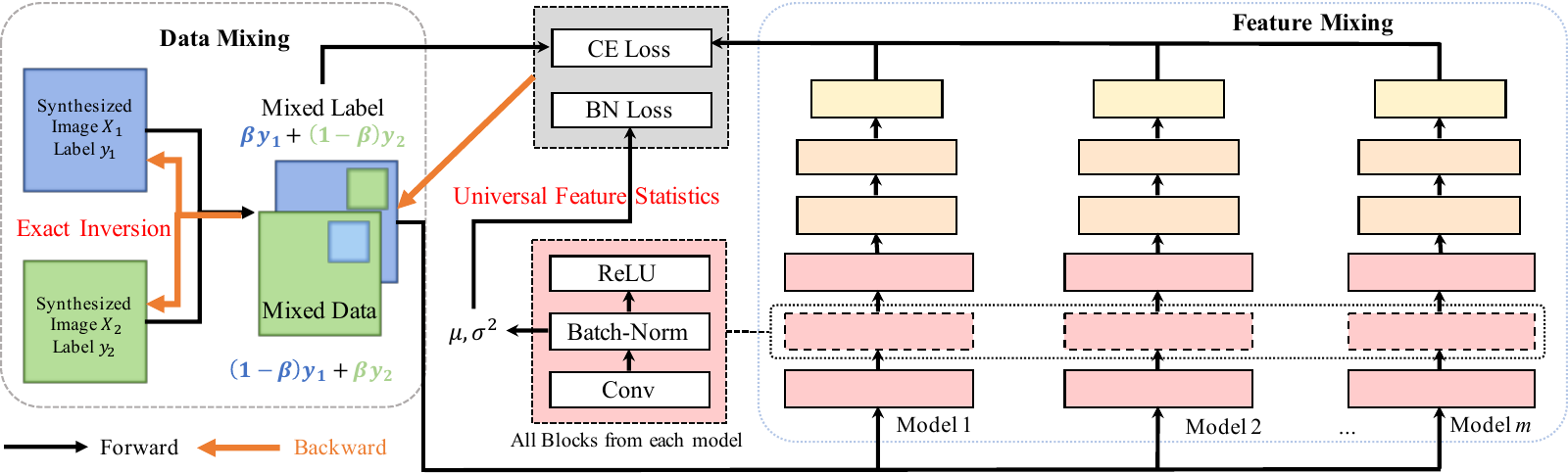}
    \caption{The overall pipeline of the proposed MixMix algorithm. Data Mixing can reduce the incorrect solution space by mixing the pixel and label of two trainable images. Feature Mixing incorporates the universal feature space from various models and generates a one-for-many synthesized dataset, after which the synthesized data can be applied to any models and application.}
    \label{fig:my_label}
\end{figure*}

However, one cannot compress the neural networks aggressively without the help of data. As an example, most full precision models can be safely quantized to 8-bit by directly rounding the parameters to their nearest integers~\cite{krishnamoorthi2018quantizing, nagel2019dfq}. However, when the bit-width goes down to 4, we have to perform quantization-aware training using the data collected from users to compensate for the accuracy loss. 
Unfortunately, due to the increasing privacy protection issue\footnote{\url{https://ec.europa.eu/info/law/law-topic/data-protection_en}}, one cannot get user data easily.
Moreover, the whole ImageNet dataset contains 1.2M images (more than 100 gigabytes), which consumes much more memory space than the model itself. Therefore, the data-free model quantization is more demanding now.

Recently, many works~\cite{cai2020zeroq, haroush2020knowledge, yin2020deepinverse} propose to \textit{invert} images from a specific pretrained model. They try to match the activations' distribution by comparing the recorded running mean and running variance in the Batch Normalization (BN)~\cite{ioffe2015bn} layer. 
Data-free model compression with the generative adversarial network is also investigated in~\cite{chen2019dafl, DBLP:journals/corr/abs-1912-11006, dfqgan}.
All these works put their focus on developing a better criterion for inverting the data from a specific model. We call this type of data as \textit{model-specific} data. 

We identify two problems of model-specific data. First, the synthesized images generated by a specific model are biased and cannot be generalized to another. 
For example, in DeepInversion~\cite{yin2020deepinverse}, synthesizing 215k 224$\times$224 resolution images from ResNet-50v1.5~\cite{res50v1.5} requires 28000 GPU hours. These images cannot be used for another model easily. As we envisioned in Fig.~\ref{fig_cross_validation}, data inverted from MobileNetV2 has 4\% higher accuracy than ResNet-50 data on MobileNetV2 quantization and vice versa. 
Therefore, model-specific data inversion requires additional thousands of GPU hours to adapt to compression on another model.
Second, due to the non-invertibility of the pretrained model, model-specific data results in inexact synthesis. A simple example is not hard to find: given a ReLU layer that has 0 in its output tensor, we cannot predict the corresponding input tensor since the ReLU layer outputs 0 for all negative input. As a result, finding the exact inverse mapping of a neural network remains a challenging task.

In this work, we propose \textit{MixMix} data synthesis algorithm that can generalize across different models and applications, which contributes to an overall improvement for data-free compression. MixMix contains two sub-algorithms, the first one is \textit{Feature Mixing} which utilizes the universal feature produced by a collection of pre-trained models. We show that Feature Mixing is equal to optimize the maximum mean discrepancy between the real and synthesized images. Therefore, the optimized data shares high fidelity and generalizability. The second algorithm is called \textit{Data Mixing} that can narrow the inversion solution space to synthesize images with exact label information.
To this end, we summarize our core contributions as:
\begin{enumerate}[nosep, leftmargin=*]
   \item \textbf{Generalizability:} We propose Feature Mixing that can absorb the knowledge from widespread pre-trained architectures. As a consequence, the synthesized data can generalize well to any models and applications.
   \item \textbf{Exact Inversion:} We propose Data Mixing which is able to prevent incorrect inversion solutions. Data Mixing can preserve correct label information.
   \item \textbf{Effectiveness:} We verify our algorithm from both theoretical and empirical perspectives. Extensive data-free compression applications, such as quantization, pruning and knowledge distillation are studied to demonstrate the effectiveness of MixMix data, achieving up to 20\% absolute accuracy boost.
\end{enumerate}

\section{Related Works}
\paragraph{Data-Driven Model Compression}
Data is an essential requirement in model compression. For example, automated exploring compact neural architectures~\cite{zoph2016nas, liu2018darts} requires data to continuously train and evaluate sub-networks. Besides the neural architecture search, quantization is also a prevalent method to compress the full precision networks. For the 8-bit case where the weight quantization nearly does not affect accuracy, only a small subset of calibration images is needed to determine the activation range in each layer, which is called Post-training Quantization~\cite{banner2019aciq, krishnamoorthi2018quantizing}. AdaRound~\cite{nagel2020adaround} learns the rounding mechanism of weights and improves post-training quantization by reconstructing each layer outputs. Quantization-aware finetuning~\cite{Esser2020LEARNED} can achieve near-to-original accuracy even when weights and activations are quantized to INT4. But this method requires a full training dataset as we mentioned. Apart from quantization, network pruning and knowledge distillation are also widely explored~\cite{hinton2015distilling, he2017channel}. 
\paragraph{Data-Free Model Compression}
The core ingredient in data-free model compression is image synthesis so that no real images are needed. Currently, the generation process can be classified into two categories, (1) directly learn images by gradient descent or (2) train a generative adversarial network (GAN) to produce images.
DAFL~\cite{chen2019dafl} and GDFQ~\cite{choi2020data} apply GAN to generate images and learn the student network. This type of work achieves good results on tiny dataset. However, training large-scale GAN requires significant efforts.
A parallel ax in image synthesis is \textit{model inversion}~\cite{mahendran2015understanding, mahendran2016visualizing}. Mordvintsev~\etal proposed DeepDream~\cite{mordvintsev2015inceptionism} to `dream' objects features onto images from a single pre-trained model. Recently, DeepInversion~\cite{yin2020deepinverse} uses the BN statistics variable as an optimization metric to distill the data and obtain high-fidelity images. BN scheme has also achieved improvements in other tasks: ZeroQ~\cite{cai2020zeroq} and \textit{the Knowledge Within}~\cite{haroush2020knowledge} use distilled dataset to perform data-free quantization, but their methods are model-specific, i.e., one generated dataset can only be used for one model's quantization. Our MixMix algorithm mainly focus on the direct optimization of data, however, it is also applicable to generative data-free application.

\section{Preliminary}
In this section, we will briefly discuss the background of how to synthesize images from a single pretrained model, then we will discuss two challenges of this method.

\subsection{Model-Specific Data Inversion}
Suppose we have a trainable image $X$ with the size of $[w, h, c]$ (in ImageNet dataset~\cite{deng2009imagenet}, the size is $224\times 224\times 3$) and a pretrained network $A:X\rightarrow Y$, the Inceptionism~\cite{mordvintsev2015inceptionism} can invert the knowledge by assigning the image with a random label $Y$, since the networks have already captured the class information. Using the cross-entropy loss, the image can be optimized by
\begin{equation}
   \min_{X} L_{\mathrm{CE}}({A}(X), Y).
\end{equation}
Recently, \cite{cai2020zeroq, haroush2020knowledge, yin2020deepinverse} observe that the pretrained networks have stored the activation statistics in the BN layers (\ie running mean and running variance). Consequently, it is reasonable for synthesized images to mimic the activation distribution of the natural images in the network. Therefore, assuming the activation (regardless of the batch) in each layer is Gaussian distributed, the BN statistics loss can be defined as 
\begin{equation}
   L_{\mathrm{BN}} = \sum_{i=1}^{\ell} (||\mu_i(X)-\hat{\mu}_i||_2+||\sigma^2_i(X)-\hat{\sigma}^2_i||_2),
\end{equation}
where $\mu_i(X)$($\sigma^2_i(X)$) is the mean (variance) of the synthesized images activation in $(i)$-th layer while $\hat{\mu}_i$ ($\hat{\sigma}^2_i$) is the stored running mean (variance) in the BN layer. Note that we can replace the MSE loss to Kullback-Leibler divergence loss as did in \cite{haroush2020knowledge}. 

In addition, an image prior loss can be imposed on $X$ to ensure the images are generally smooth. In~\cite{haroush2020knowledge}, the prior loss is defined as the MSE between $X$ and its Gaussian blurred version $\varepsilon(X)$. In this work, we use the prior loss defined in~\cite{yin2020deepinverse}: $L_{\mathrm{prior}}(X) = \mathcal{R}_{\mathrm{TV}}(X)+\lambda_{\mathrm{\ell_2}}\mathcal{R}_{\mathrm{\ell_2}}(X)$, which is the sum of variance and norm regularization. Combining these three losses, the final minimization objective of knowledge inversion for a specific model can be formulated as:
\begin{equation}
   \min_{X}\ \lambda_1L_{\mathrm{CE}}(X)+\lambda_2L_{\mathrm{BN}}(X)+\lambda_3L_{\mathrm{prior}}(X)
   \label{eq_sin_teacher}
\end{equation}

\subsection{Biased Feature Statistics}
\label{sec_drawbacks}

\begin{table}[t]
   \caption{BN Loss comparison on ResNet-50 and MobileNetV2, given different types of data.}
   \centering
   \begin{adjustbox}{max width=\linewidth}
   \begin{tabular}{lccc}
   \toprule 
   {{Model}} & {ImageNet} & {Res50 Data} & {MobV2 Data} \\
   \midrule
   ResNet-50 & 0.018 & 0.049 & 0.144 \\
   MobileNetV2 & 0.722 & 4.927 & 1.498 \\
   \bottomrule
   \end{tabular}
\label{tab_bnloss}
\end{adjustbox}
\end{table}

For image synthesis tasks, the real ImageNet dataset could be viewed as the global minimum, which can be utilized to perform model compression on any neural architectures. However, we find that the data synthesized from one model cannot be directly applied to another different architecture. Example results demonstrated in Fig.~\ref{fig_cross_validation} show that data synthesized on ResNet-50 gets bad quantization results on 
MobileNetV2 (4\% lower) and vice versa. 

We conjecture the reason for this phenomenon is the different feature statistics learned in different CNNs. 
In each neural network, the distribution characteristics of the training data are encoded and processed in its unique feature space (\ie unique BN statistics). Therefore, extract the feature information from this neural network leads to biased statistics. To test this, we train images from ResNet-50 and MobileNetV2 by Eq.~(\ref{eq_sin_teacher}) and validate the synthesized images on these architectures. Results in Table~\ref{tab_bnloss} show that $L_{\mathrm{BN}}$ of the MobileNetV2 data is high when evaluated on ResNet50. Interestingly, in an opposite way, ResNet-50 data also has poor performance on MobileNetV2. However, as shown in Table~\ref{tab_bnloss}, $L_{\mathrm{BN}}$ always remains in low-level regardless of model in the case of real ImageNet data. 

\subsection{Inexact Inversion}
\label{sec_non_invert}
\begin{figure}[t]
  \begin{center}
  \includegraphics[width=\linewidth]{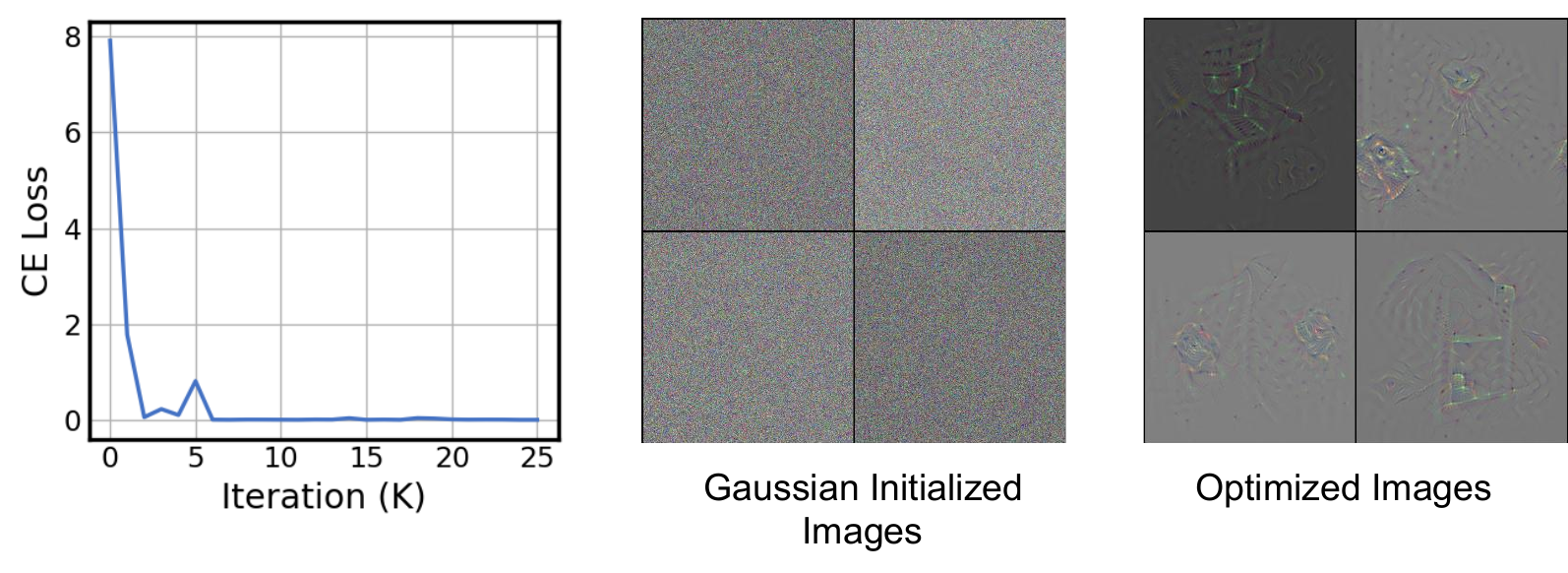}
  \end{center}
  \caption{Optimizing cross-entropy loss of the images can be very fast and down to 0, but the images nearly have no class features.}
  \label{fig_ce}
\vspace{-3mm}
\end{figure}

Apart from BN statistics loss $L_{\mathrm{BN}}(X)$, the cross-entropy loss $L_{\mathrm{CE}}(X)$ desires to learn an inverted mapping from label to input images, denoted by $A^{-1}:Y\rightarrow X$.
According to~\cite{behrmann2019invertible}, a residual block is invertible if it has less than 1 Lipschitz constant and the same input-output dimension (\ie $\mathbb{R}^d\rightarrow\mathbb{R}^d$). We find the latter one is not true for all classification models, since the label dimension is $Y\in[0, 1]^{1000}$ and input dimension is $X\in\mathbb{R}^{224\times 224\times 3}$ for the ImageNet classification task. This dimension difference can produce a huge solution space for image inversion.
Let us consider an example of the average pooling layers:
\begin{example}
\label{example}
Consider a $2\times 2$ AvgPool layer $o=W^TV$, where $V\in\mathbb{R}^4$ and $W=[0.25, 0.25, 0.25, 0.25]$ are the input and weight vectors. This AvgPool layers is non-invertible because given output $o$, any inputs that has same mean as $o$ will suffice the condition. 
\end{example}
Therefore, finding the exact input is infeasible in such a large space.
In fact, almost every CNN has a $7\times 7$ AvgPool layer before the final fully-connected layer.
To visualize this, we optimize 4 images using $L_{\mathrm{CE}}(X)$ and $L_{\mathrm{prior}}(X)$ and plot the training curve in Fig.~\ref{fig_ce}. It is clear that CE loss is easy to optimize, yet we cannot invert real images that contain rich class information.

\section{Methodology}
In this section, we introduce the proposed \textit{MixMix} algorithm, which can improve both the generalizability and the invertibility of the dataset. 

\subsection{Feature Mixing}
Before delving into the proposed algorithm, we would like to discuss \textit{the general problem of comparing samples from two probability distributions}. This problem is of great interest in many areas such as bioinformatics where two image types are evaluated to determine if they come from the same tissues. In the case of image synthesis, we also need to evaluate if the generated images preserve high fidelity.
Formally, given two distributions $p$ and $q$ with their random variables defined on a topological space $\mathcal{X}$ and observations ($X=\{x_1,\dots, x_m\}, Z=\{z_1,\dots, z_n\}$) that are independently and identically distributed (i.i.d.) from $p$ and $q$, can we find some methods to determine if $p=q$? In~\cite[Lemma 9.3.2]{dudley2018real}, this problem is well-defined as 
\begin{lemma}[{{\cite{dudley2018real}}}]
   Let $(\mathcal{X},d)$ be a metric space, and let $p, q$ be two Borel probability measures defined on $\mathcal{X}$. Then $p = q$ if and only if $E_{x\sim p}(f(x)) = E_{z\sim q}(f(z))$ for all $f \in C(\mathcal{X})$, where $C(\mathcal{X})$ is the space of bounded continuous functions on $\mathcal{X}$.
\end{lemma}
However, evaluating all $f\in C(\mathcal{X})$ in the finite settings is not practical. Fortunately, Gretton \etal~\cite{gretton2012kernel} proposed the maximum mean discrepancy that can be represented by:
\begin{equation}
   \text{MMD}[\mathcal{F}, X, Z] = \sup_{f\in \mathcal{F}}\left(E_{x\sim p}[f(x)] - E_{z\sim q}[f(z)]\right),
\end{equation}
where $\mathcal{F}$ is a class of function which is much smaller than $C(\mathcal{X})$. 
We will directly give two results that are related to MMD theory and the BN statistics loss. The detailed derivation is located in the appendix. 
\begin{enumerate}[nosep, leftmargin=*]
   \item Let $\mathcal{H}$ be a \textit{reproducing kernel Hilbert space (RKHS)}, and when we set $\mathcal{F}$ to $\mathcal{H}$, the squared MMD is given by.
   \begin{equation}
      \text{MMD}^2[\mathcal{H}, X, Z] = ||\mu_p-\mu_q||^2_{\mathcal{H}},
   \end{equation}
   where $\mu_p\in\mathcal{H}$ is called the mean embedding of $p$. we can use the feature extractor of the CNNs and its feature-map to define a reproducing kernel. Therefore the running mean and variance\footnote{The variance can be defined in the kernel's second-order space.} in BN layers can be treat to the mean embedding of $\mathcal{H}$. \textit{Thus optimizing the $L_{\mathrm{BN}}$ is equivalent to minimize the $\text{MMD}^2$ between the real images and the synthesized images.}
   \item \cite[Theorem 5]{gretton2012kernel} states that the kernel for feature extraction $k$ must be universal\footnote{The universal $\mathcal{H}$ means that, for any given $\epsilon>0$ and $f\in C(\mathcal{X})$ there exists a $g\in \mathcal{H}$ such that the max norm $||f-g||_{\infty}<\epsilon$.} so that we have $\text{MMD}^2[\mathcal{H}, X, Z]=0$ if and only if $p=q$.
\end{enumerate}

\noindent These two results indicate that if the neural network is universal and the $L_{\mathrm{BN}}$ of synthesized images is 0, we could say the generated images are very close to real images. 
But it turns out few neural networks are universal and are only possible for extremely deep or wide ones~\cite{lin2017does, lu2017expressive}. In Sec.~\ref{sec_drawbacks}, we show that different CNN has different feature statistics. And the data generated from one model is hard to be transferred to another one. This empirical evidence further illustrates the lack of universality in one model. 
We hereby propose Feature Mixing, which gathers the knowledge in the model zoo and aims to average out the feature bias produced by each individual. We expect the aggregation of pre-trained models can improve the universality of their corresponding RKHS.
To demonstrate that mixing features can improve the universality, we have the following theorem:
\begin{theorem}
   \label{thm}
   Assume there are $m$ neural networks with ReLU activation function 
   ($A_i:X^d\rightarrow \mathbb{R}$). Then, the averaged model $\frac{1}{m}\sum_{i=1}^mA_i$ is universal if $m\ge \mathrm{ceil}(\frac{d+1}{w})$, where $w$ is the averaged width.
\end{theorem}
Proof is available in the appendix. Although sufficing the ultimate universality may require a large $m$, we anticipate increasing the number of features mixed can improve the quality as well as the generalizability of the generated data.
To this end, we will show how to apply Feature Mixing during synthesis. Consider a pretrained model zoo $\mathcal{M}=\{A_1, A_2, \dots, A_m\}$, the Feature Mixing aims to optimize:
\begin{equation}
   \min_{X} \frac{1}{m}\sum_{i=1}^{m}\left(\lambda_{1}L_{\mathrm{CE}}(X, A_i)+\lambda_{2}L_{\mathrm{BN}}(X, A_i)\right).
   \label{eq_fmix}
\end{equation}
Note that we also add a prior loss on $X$. However, assuming the identical size of each model $A_i$, the training memory and computation is linearly scaled by the number of features we mix. Therefore, we add a hyper-parameter $m^{\prime}\le m$, which will decide how many models will be sampled from the model zoo for each batch of the data. 
The effect of $m^{\prime}$ will be studied in the experiments. 
Another problem is how to select model zoo, we expect different architecture families will contain different feature statistics, therefore we choose model families as many as possible in our model zoo.

\subsection{Data Mixing}

\begin{figure}[t]
   \begin{center}
   \includegraphics[width=\linewidth]{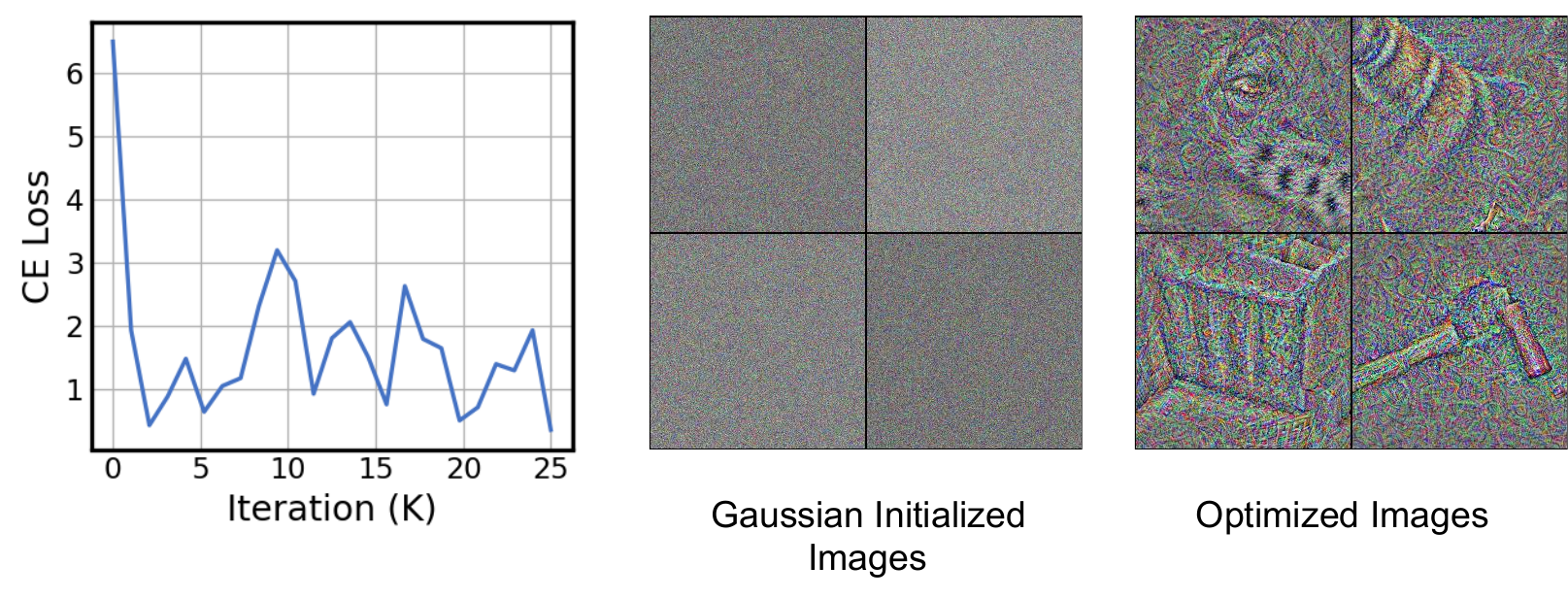}
   \end{center}
   \caption{Training images with Data Mixing.}
   \label{fig_ce_mix}
   \vspace{-3mm}
 \end{figure}

Another problem in image synthesis is that certain layers or blocks in neural network cause inexact inversion, as we described in Sec.~\ref{sec_non_invert}. In this section, we show that we can mitigate this problem through Data Mixing. Denote two image-label pairs as $(X_1, Y_1)$ and $(X_2, Y_2)$, we first randomly generate a binary mask $\alpha\in\{0,1\}^{w\times h}$. The elements in this mask will be set to 1 if it is located in the bounding box: 
\begin{equation}
   \alpha_{ij} = \begin{cases}
      1& \text{if } x_l\le i\le x_r \text{ and } y_d\le j\le y_u\\
      0& \text{otherwise}
      \end{cases},
\end{equation}
where $x_l, x_r, y_d, y_u$ are the left, right, down, up boundaries of the box. The center coordinate of the bounding box can be computed by $(\frac{x_l+x_r}{2}, \frac{y_d+y_u}{2})$. With this binary mask, we can mix the feature of data, given by
\begin{equation}
   \hat{X} = (1-\alpha) X_1 + \alpha g(X_2).
\end{equation}
Here $g(\cdot)$ is a linear interpolation function that can resize the images to the same size of the bounding box. The mixed data $\hat{X}$ now contains information from two images, thus we mix the label so that the CE loss becomes 
\begin{equation}
   L_{\mathrm{CE}}(\hat{X}) = (1-\beta) L_{\mathrm{CE}}(\hat{X}, Y_1) + \beta L_{\mathrm{CE}}(\hat{X}, Y_2) 
\end{equation}
where $\beta$ is computed as the ratio of bounding box area and the image area $w\times h$. Data Mixing is inspired by the success of mixing-based data-augmentations, such as CutMix~\cite{yun2019cutmix} and Mixup~\cite{zhang2017mixup}. They use mixed data and label to train a model that has stronger discriminative power. 
Indeed, such augmentation techniques is also helpful in our scope by generating the robust features that remain discriminative when being mixed.
Moreover, in this work, data mixing is also used to reduce the inexact inversion solutions of the neural networks. Back to Example~\ref{example}, for each iteration $t$, we must suffice $\mathrm{mean}(\hat{V}^t)=\hat{o}^t$, where the input and output are mixed differently and therefore more restrictions are added when inverting the input. We also give an example to illustrate this:
\begin{example}
   Consider the same $2\times 2$ AvgPool layer in former discussion. If  $o=1$, then we have
   \begin{equation}
      V_1 + V_2 + V_3 + V_4 = 4.
   \label{eq_nomix}
   \end{equation}
   Now assume a mixed input $\hat{V}=
   [\hat{V}_1, \hat{V}_2, V_3, V_4]$ and output $\hat{o}=0$ where the first two elements in $\hat{V}$ come from another input image. Then we can obtain following relationships:
   \begin{equation}
   \begin{cases}
      V_1 + V_2 + V_3 + V_4 = 4 \\
      \hat{V}_1 + \hat{V}_2 + V_3 + V_4 = 0
   \end{cases}
   = 
   \begin{cases}
      V_3+V_4=4 \\
      V_1+V_2 = 0 \\
      \hat{V}_1+\hat{V}_2 = -4
   \end{cases}
   \label{eq_datamix}
   \end{equation}
\end{example}
We can see that Data Mixing can help image inversion because the solution space in Eq.~(\ref{eq_datamix}) is much smaller than Eq.~(\ref{eq_nomix}).
We also visualize the data-feature mixing using only CE and prior loss for image generation. The training curve as well as the optimized images are shown in Fig.~\ref{fig_ce_mix}.
Notably, there are some basic shapes or textures in optimized images if we mix the data features.

Together with Feature Mixing, we formalize the MixMix data synthesis procedure in algorithm \ref{alg:1}.

\begin{algorithm}[t]
       \caption{MixMix data synthesis}
       \label{alg:1}
       \KwIn{Pretrained model zoo, subset size $m^{\prime}$}
      Initialize random label $Y_i$\;
      Randomly select $m^{\prime}$ models\;
      \For{all $t=1,2,\dots,T$-iteration}
      {
         Generate mask and mix data features\;
         \For{all $j=1,2,\dots,m^{\prime}$-th pretrained model ${A}_j$}
         {
               Compute BN statistic loss $L_{\mathrm{BN}}(\hat{X}, \hat{\mu}_j, \hat{\sigma}_j)$\;
               Compute mixed CE loss $L_{\mathrm{CE}}({A}_j(\hat{X}))$\;
         }
         Compute image prior loss $L_{\mathrm{prior}}(X_i)$\;
         Descend final loss objective and update $X$
      }
      \textbf{return} MixMix data $X$
\end{algorithm}

\section{Experiments}

We conduct our experiments on both CIFAR10 and ImageNet dataset. 
We set the subset size of Feature Mixing as $m^{\prime}=3$ except we mention it. We select 21 models in our pretrained model zoo, including ResNet, RegNet, MobileNetV2, MobileNetV3, MNasNet, VGG, SE-Net, DenseNet, ShuffleNetV2~\cite{he2016resnet, radosavovic2020regnet, sandler2018mobilenetv2, howard2019searching, tan2019mnasnet, simonyan2014vgg, ma2018shufflenet, hu2018senet, huang2017densenet}, \etc See detailed descriptions in appendix.
The width and the height of the bounding box for data mixing are sampled from a uniform distribution. We use Adam~\cite{kingma2014adam} optimizer to optimize the images. Most hyper-parameters and implementation are aligned up with \cite{yin2020deepinverse}, such as multi-resolution training pipeline and image clip after the update. We optimize the images for 5k iteration and use a learning rate of 0.25 followed by a cosine decay schedule. To determine $\lambda$, we set it learnable and optimize it by gradient descent, details can be found in the appendix. Training 1024 MixMix images requires approximately 2 hours on 8 1080TIs.

\subsection{Analysis of the Synthesized Images}
\begin{figure}[t]
   \centering
   \includegraphics[width=\linewidth]{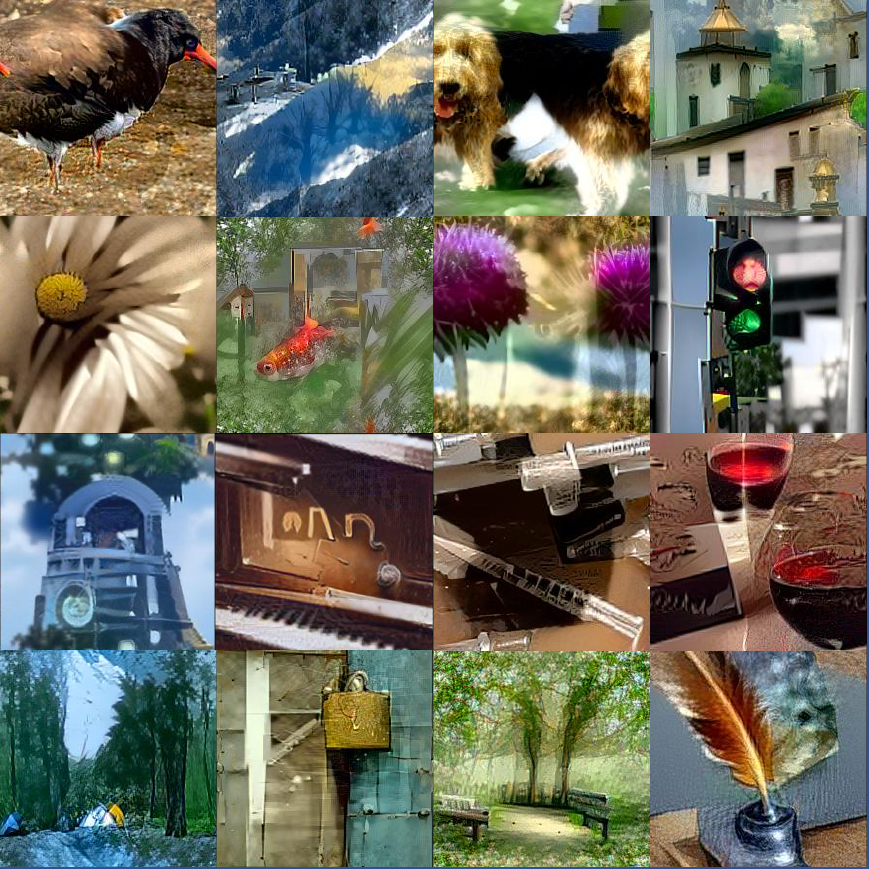}
   \caption{Example images synthesized by 21 different models (Labels: backpack, overskirt, stingray, hen, gondola, anemone, white stork, Norwich terrier,
   husky, canoe, banister, crock pot, beaker, langur, church, lighthouse).}
   \label{fig_syn_images}
\end{figure}

We present some qualitative evaluation in Fig.~\ref{fig_syn_images}. Notably, MixMix data preserves high fidelity and resembles real images. 
To test the generalizability of the synthesized images, we report the average classification accuracy (as well as the standard deviation) on 21 different models in the model zoo. We also report the Inception Score (IS)~\cite{NIPS2016_8a3363ab} to evaluate the image quality. 
\begin{table}[htbp]
\caption{Classification accuracy evaluated on 21 different models and the Inception Score (IS) metric of the synthesized images.}
\centering
\begin{adjustbox}{max width=\linewidth}
\begin{tabular}{lrrr}
\toprule
\textbf{Methods} & \textbf{Size} & \textbf{Average Acc.} & \textbf{IS}\\
\midrule
DeepDream-R50~\cite{mordvintsev2015inceptionism} & 224 & 24.9$\pm$8.23 & 6.2\\
DeepInversion-R50~\cite{yin2020deepinverse} & 224 & 85.96$\pm$5.80 & 60.6 \\
\midrule
BigGAN~\cite{brock2018bigGAN} & 256 & N/A & 178.0 \\
SAGAN~\cite{zhang2019SAGAN} & 128 & N/A & 52.5 \\
\midrule
MixMix & 224 & \textbf{96.95$\pm$1.53} & 92.9 \\
\bottomrule
\end{tabular}
\end{adjustbox}
\label{tab_gen_div}
\end{table}

It can be noted from Table~\ref{tab_gen_div} that model-specific data (from ResNet-50) has lower average accuracy. The proposed MixMix data achieves near 97\% average accuracy and the stablest result. This means that our generated images have obvious class characteristics that all models can recognize. As a consequence, we can safely adopt it for all data-free applications and any architecture. We also compare with some GAN-based image synthesis methods where MixMix can achieve a comparable Inception Score. 

\subsection{Data-Free Quantization}
In this section, we utilize the images generated by MixMix to conduct Post-Training Quantization (PTQ) and Quantization-Aware Training (QAT) on ImageNet. Here we adopt two state-of-the-art PTQ and QAT methods, namely BRECQ~\cite{li2021brecq} and LSQ~\cite{Esser2020LEARNED}.

\paragraph{Post-Training Quantization}
\cite{li2021brecq} uses the block output reconstruction to optimize the rounding policy of weight quantization and the activation quantization range. 
We use 1024 images with a batch size of 32 to optimize the quantized model. Each block is optimized by 20k iterations. 
We compare ZeroQ~\cite{cai2020zeroq} and \textit{The Knowledge Within}~\cite{haroush2020knowledge} (abbreviated as KW in the this paper) data on ResNet-50, MobileNet-b (modified version of MobileNetV1~\cite{howard2017mobilenets} without BN and ReLU after the depthwise convolutional layers), MobileNetV2, and MNasNet. To fairly compare results, we implement ZeroQ and KW to the same pre-trained models and apply the same quantization algorithm. Results are demonstrated in Table~\ref{tab_ptq}.
For ResNet-50 4-bit quantization, MixMix data has comparable results with the real images (only 0.14\% accuracy drop). When the bit-width of weight goes down to 2, the quantization becomes more aggressive and requires high-fidelity data for calibration. In that case, MixMix still achieves the lowest degradation among the existing methods. We next verify data-free quantization on three mobile-platform networks, which face a higher risk of performance degeneration in compression. 
It can be seen that 4-bit quantization on lightweight networks generally will have much higher accuracy loss even using real images. However, MixMix data still reaches close-to-origin results. As an example, ZeroQ and KW only reach 49.83\% and 59.81\% accuracy on MobileNetV2, respectively. Whereas, MixMix can boost the performance to 64.01\%.

\begin{table}[t]
   \caption{ImageNet top-1 accuracy comparison on PTQ.}
   \centering
   \begin{adjustbox}{max width=\linewidth}
   \begin{tabular}{lllr}
   \toprule 
   \multirow{1}{4em}{\textbf{Model}} & \textbf{Bits(W/A)} & \textbf{Data Source} & \textbf{Top-1 Acc.} \\
   \midrule
   \multirow{8}{4.5em}{ResNet-50\\FP: 77.00} & 4 / 4 & Training data & 74.72 \\
   & 4 / 4 & ZeroQ~\cite{cai2020zeroq} & 73.73 \\
   & 4 / 4 & KW~\cite{haroush2020knowledge} & 74.05 \\
   & 4 / 4 & MixMix & \textbf{74.58} \\
   \cmidrule(l{2pt}r{2pt}){2-4}  
   & 2 / 4 & Training data & 68.87\\
   & 2 / 4 & ZeroQ~\cite{cai2020zeroq} & 64.16 \\
   & 2 / 4 & KW~\cite{haroush2020knowledge} & 57.74 \\
   & 2 / 4 & MixMix & \textbf{66.49} \\
   \midrule
   \multirow{4}{5.5em}{MobileNet-b\\FP: 74.53} & 4 / 4 & Training data & 66.11 \\
   & 4 / 4 & ZeroQ~\cite{cai2020zeroq} & 55.93 \\
   & 4 / 4 & KW~\cite{haroush2020knowledge} & 61.94 \\
   & 4 / 4 & MixMix & \textbf{65.38} \\
   \midrule
   \multirow{4}{5.5em}{MobileNetV2\\FP: 72.49} & 4 / 4 & Training data & 64.61 \\
   & 4 / 4 & ZeroQ~\cite{cai2020zeroq} & 49.83 \\
   & 4 / 4 & KW~\cite{haroush2020knowledge} & 59.81 \\
   & 4 / 4 & MixMix & \textbf{64.01} \\
   \midrule
   \multirow{4}{4.5em}{MNasNet\\FP: 73.52} & 4 / 4 & Training data & 58.86 \\
   & 4 / 4 & ZeroQ~\cite{cai2020zeroq} & 52.04 \\
   & 4 / 4 & KW~\cite{haroush2020knowledge} & 55.48 \\
   & 4 / 4 & MixMix & \textbf{57.87} \\
   \bottomrule
   \end{tabular}
\label{tab_ptq}
\end{adjustbox}
\end{table}

\paragraph{Quantization-Aware Training}
QAT aims to recover the performance of the quantization neural network in low-bit scenarios. In this work, we leverage the state-of-the-art QAT baseline: Learned Step Size Quantization~\cite{Esser2020LEARNED}. In QAT, Straight-Through Estimator (STE) is adopted to compute the gradients of the latent weights. In the case of LSQ, STE is also used to estimate the gradients of quantization step size. Note that our QAT uses per-layer quantization which is more challenging than per-channel quantization used in PTQ.
We synthesize 100k images and use a batch size of 128 to finetune the quantization neural network. During QAT, the full precision model serves as the teacher and we use KL loss with temperature $\tau=3$ as the criterion. We also incorporate the intermediate feature loss proposed in~\cite{haroush2020knowledge}. We finetune the quantized model for 44000 steps and it only takes 3 hours on 8 1080TIs to complete the finetuning. 
We perform W4A4 quantization-aware training on ResNet-50 and MobileNetV2. Additionally, W2A4 quantization is applied to ResNet-50. Results are presented in Table~\ref{tab_qat}. Note that the real training dataset only contains 100k images. In W4A4, we show the MixMix dataset only has 2.6\% accuracy reduction compared to the natural images. In W2A4 quantization, the gap between synthesized images and natural images is bigger. We compare an existing work, KW~\cite{haroush2020knowledge}, where our MixMix can recover 1.7\% accuracy in 4-bit MobileNetV2.

\begin{table}[t]
   \caption{ImageNet top-1 accuracy comparison on QAT.}
   \centering
   \begin{adjustbox}{max width=\linewidth}
   \begin{tabular}{lllr}
   \toprule 
   \multirow{1}{4em}{\textbf{Model}} & \textbf{Bits(W/A)} & \textbf{Data Source} & \textbf{Top-1 Acc.} \\
   \midrule
   \multirow{4}{4.5em}{ResNet-50\\FP:77.00} & 4 / 4 & Training data & 76.09 \\
   & 4 / 4 & MixMix & \textbf{73.39} \\
   \cmidrule(l{2pt}r{2pt}){2-4}  
   & 2 / 4 & Training data & 70.20\\
   & 2 / 4 & MixMix & \textbf{64.60} \\
   \midrule
   \multirow{3}{5em}{MobileNetV2\\FP:72.49} & 4 / 4 & Training data & 68.50 \\
   & 4 / 4 & KW~\cite{haroush2020knowledge} & 66.07 \\
   & 4 / 4 & MixMix & \textbf{67.74} \\
   \bottomrule
   \end{tabular}
\label{tab_qat}
\end{adjustbox}
\end{table}

\subsection{Data-Free Pruning}
\begin{table}[t]
   \caption{ImageNet top-1 accuracy comparison on pruning. UP and SP refer to unstructured and structured pruning, respectively.}
   \centering
   \begin{adjustbox}{max width=\linewidth}
   \begin{tabular}{lllr}
   \toprule 
   \multirow{1}{4em}{\textbf{Model}} &  \textbf{Data Source} & \textbf{UP Acc.} & \textbf{SP Acc.}\\
   \midrule
   \multirow{3}{4.5em}{ResNet-50\\FP: 77.00} & Training data & 76.53 & 70.95 \\
    & DeepInversion~\cite{yin2020deepinverse} & 71.58 & 65.07 \\
    & MixMix & \textbf{75.41} & \textbf{69.80} \\
    \midrule
    \multirow{3}{5.5em}{MobileNet-b\\FP: 74.53} & Training data & 72.96 & 48.38 \\
    & DeepInversion~\cite{yin2020deepinverse} & 70.56 & 40.62 \\
    & MixMix & \textbf{70.64} & \textbf{44.82} \\
   \midrule
   \multirow{3}{5.5em}{MobileNetV2\\FP: 72.49} & Training data & 68.96 & 45.24 \\
   & DeepInversion~\cite{yin2020deepinverse} & 47.08 & 15.32 \\
   & MixMix & \textbf{66.74} & \textbf{42.47} \\
   \midrule
   \multirow{3}{4.5em}{MNasNet\\FP: 73.52}& Training data & 70.43 & 47.81  \\
   & DeepInversion~\cite{yin2020deepinverse} & 57.42 & 22.62 \\
    & MixMix & \textbf{67.98} & \textbf{43.41}\\
   \bottomrule
   \end{tabular}
\label{tab_prune}
\end{adjustbox}
\end{table}

Another important compression technique is network pruning. In this section, we validate both unstructured pruning~\cite{han2015deepcompress} and structured channel pruning~\cite{he2017channel} using L1-norm magnitude measure. The pruning ratio (or sparsity) is set to 0.5 and 0.2 for unstructured and structured pruning. We also use 1024 images to reconstruct the output feature-maps after pruning as did in \cite{he2017channel}. We mainly compare DeepInversion~\cite{yin2020deepinverse} as the baseline method. 
Table~\ref{tab_prune} summarizes the results, where the baseline still produces a large gap between the real images. Quantitatively, on sparse ResNet-50, MixMix is nearly 4\% higher than DeepInversion. We find the hard case for pruning is MobileNetV2 and MNasNet, especially when conducting channel pruning.
In these cases, the improvement of MixMix is much more evident, up to 27\% absolute accuracy uplift on MobileNetV2 and 21\% on MNasNet.

\begin{table*}[t]
   \caption{Generalizability study via cross-validation. We apply post-training quantization to target models and quantized them to W4A4 (except for EfficientNet which is W4A8). MixMix data only requires one-time synthesis and is generaliable to all models.}
   \centering
   \begin{adjustbox}{max width=0.95\linewidth}
   \begin{tabular}{|l|c|c|c|c|c|c|}
   \hline
   \diagbox{Source Data}{Target Model} & ResNet-18* & ResNet-50 & MobileNet-b* & MobileNetV2 & MNasNet & EfficientB0* \\
   \hline
   KW-ResNet-18 & 69.08 & 73.84 & 61.06 & 59.79 & 53.12 & 67.67\\
   KW-ResNet-50 & 67.34 & 74.05 & 61.07 & 57.31 & 47.57 & 68.33\\
   KW-MobileNet-b & 68.28 & 72.83 & 61.95 & 53.04 & 50.02 & 58.84\\
   KW-MobileNetV2 & 63.58 & 70.20 & 60.81 & 59.81 & 54.08 & 68.59\\
   KW-MNasNet & 66.03 & 71.45 & 54.86 & 59.03 & 55.48 & 68.15\\
   KW-EfficientNetB0 & 65.87 & 71.34 & 46.28 & 60.47 & 42.84 & 69.59\\
   MixMix & \textbf{69.46} & \textbf{74.58} & \textbf{65.38} & \textbf{64.01} & \textbf{57.87} & \textbf{70.59} \\
   Training Set & 69.52 & 74.72 & 66.11 & 64.63 & 58.86 & 70.64 \\
   \hline
   \multicolumn{7}{l}{*These models are exceluded from our pretrained model zoo, thus they can test the generalizability of MixMix data.}
   \end{tabular}
\label{tab_gen}
\end{adjustbox}
\end{table*}

\subsection{Data-Free Knowledge Distillation}
In this section, we perform knowledge distillation to verify MixMix. Due to the reproducibility issue, we retrieve the code of DAFL~\cite{chen2019dafl} as our code-base to conduct experiments on CIFAR10. 
We add our MixMix training objective during the GAN training. See the detailed description of the KD experiment settings in the appendix. Here we compare DeepInversion~\cite{yin2020deepinverse} and DAFL~\cite{chen2019dafl} as our baseline methods. The results are shown in Table~\ref{tab_kd}. We can see that our MixMix is 3\% higher than DeepInversion when distilling VGG11 to learn VGG11. Compared with the original DAFL, our method is also 2.5\% higher.

\begin{table}[t]
   \caption{CIFAR10 Data-Free Knowledge Distillation comparison.}
   \centering
   \begin{adjustbox}{max width=\linewidth}
   \begin{tabular}{lccc}
   \toprule 
   Teacher & VGG11 & VGG11 & ResNet-34 \\
   Student & VGG11 & ResNet-18 & ResNet-18 \\
   Teacher Acc. & 94.31  & 94.31 & 95.42 \\
   \midrule
   DeepInversion & 90.78 & 90.36 & 93.26 \\
   DAFL (original) & - & - & 92.22 \\
   DAFL (+ MixMix) & \textbf{93.97} & \textbf{91.57} & \textbf{94.79}\\
   \bottomrule
   \end{tabular}
\label{tab_kd}
\end{adjustbox}
\end{table}

\subsection{Generalizability Study}
In this section, we verify the generalizability and the transferability of the MixMix data. We consider it as an essential property for the optimal synthesized data, since the real images perform well whatever the model is. We conduct \textit{cross-validation}, \ie, using data inverted from a single model to validate on multiple model compression. We conduct 4-bit post-training quantization and use 1024 synthesized images. Target test models include ResNet-18, ResNet-50, MobileNet-b, MobileNet-V2, MNasNet and EfficientNetB0. Note that \textit{ResNet-18, MobileNet-b, EfficientB0 are not utilized in our pre-trained model zoo, thus their compression performances are effective evaluations of data generalizability}. As a baseline, we implement KW\cite{haroush2020knowledge} to generate images for the tested models. Thus it sufficiently extracts the information inside the model that is ready to compress. 

The results are summarized in Table~\ref{tab_gen}. A general rule for model-specific data is that its good performance is restricted to the original model compression. Take ResNet-50 as an example, synthesized data from the original model achieves 74.05\% performances, which is higher than data synthesized from other models. However, ResNet-50 data is yet 2.6\% lower than ResNet-18 data on ResNet-18 quantization. Evidence from this table proves our conjecture that the feature bias cannot be eliminated when synthesizing from one specific model. In contrast, MixMix data preserves high generalizability and is applicable to all models even when they are not used in the pre-trained model zoo. 

\subsection{Ablation Study}
In this section, we study the design choice of Feature Mixing (denoted as FMix) and Data Mixing (denoted as DMix). We verify the number of features we mixed and the usage of DMix during image synthesis. We test it on ResNet-50 structured pruning. Table~\ref{tab_ablation} presents the results of ablation experiments. We can find that large $m^{\prime}$ is helpful to increase the image quality for compression. However, we did not obtain significant accuracy improvement when $m^{\prime}$ is higher than 3. 
As for DMix, we find it can consistently improve the performance of image synthesis. However, when $m^{\prime}=4$, the improvement is also trivial. This result can be improved by more hyper-parameter tuning. Nevertheless, the effectiveness of MixMix is still evident, and in this work, we mainly use $m^{\prime}=3$ with DMix to synthesize data by taking the synthesis time into consideration.

\begin{table}[t]
   \caption{ImageNet pruning on ResNet-50, given different policies when generating the data.}
   \centering
   \begin{adjustbox}{max width=\linewidth}
   \begin{tabular}{lllr}
   \toprule 
   \multirow{1}{4em}{\textbf{Method}} &  \textbf{Acc.} & \textbf{Method} & \textbf{Acc.}\\
   \midrule
   FMix ($m^{\prime}=1$) & 64.46 & FMix ($m^{\prime}=1$)+DMix & 68.16 \\
   FMix ($m^{\prime}=2$) & 68.87 & FMix ($m^{\prime}=2$)+DMix & 69.29 \\
   FMix ($m^{\prime}=3$) & 69.49 & FMix ($m^{\prime}=3$)+DMix & 69.80 \\
   FMix ($m^{\prime}=4$) & 69.54 & FMix ($m^{\prime}=4$)+DMix & 69.47 \\
   \bottomrule
   \end{tabular}
\label{tab_ablation}
\end{adjustbox}
\end{table}

\vspace{-2mm}
\section{Conclusion}
\vspace{-2mm}

In this work, we identify two drawbacks in model-specific inversion method, namely insufficient generalizability and inexact inversion process. The proposed MixMix algorithm improves the existing method by leveraging the knowledge of a collection of models and data mixing. MixMix algorithm is both effective and efficient as it only requires one-time synthesis to generalize any models. Experimental results demonstrate that MixMix establishes a new state-of-the-art for data-free compression.

\subsection*{Acknowledgments}
This work is supproted by NSFC GP 61876032.

\nocite{li2020efficient,zhang2021diversifying,li2021mqbench,shen2019searching}
%


{\small
\bibliographystyle{ieee_fullname}
\bibliography{egbib}
}

\end{document}


\newcommand{\mix}{MixMix}

\title{Append for MixMix}

\maketitle
\ificcvfinal\thispagestyle{empty}\fi

\nocite{shen2019searching}
%


\newpage
\appendix

\section{Additional Derivation and Main Proofs}
In this section, we first recap the results from Gretton \etal~\cite{gretton2012kernel}, then we prove Theorem 4.2. From~\cite[Theorem 5]{gretton2012kernel}, as known that when the RKHS is universal, we have $p=q$ if and only if $||\mu_p-\mu_q||_{\mathcal{H}}^2 = 0$. The proof is illustrated as follows. First, $p=q$ implies MMD$^2[\mathcal{H}, X, Z]=0$. Then, we only have to prove the converse. When $\mathcal{H}$ is universal, for any given $\varepsilon>0$ and $f\in C(\mathcal{X})$, there exists a $g\in\mathcal{H}$ such that $||f-g||_{\infty} < \varepsilon$.
Now, we simplify the notion of expectation $E_{x\sim p}, E_{z\sim q}$ to $E_x, E_z$ and make an expansion that
\begin{align}
   |E_xf(x) - E_zf(z)| \le |E_xf(x) - E_xg(x)| + \nonumber\\ 
   |E_xg(x) - E_zg(z)| + |E_zg(z) - E_zf(z)|
\end{align}
By the universality of the RHKS, there exist a $g\in\mathcal{H}$ so that the first and the third term satisfy
\begin{equation}
   |E_xf(x) - E_xg(x)| \le E_x |f(x)-g(x)| \le \varepsilon
\end{equation}
For the second term, since $g\in\mathcal{H}$, $|E_xg(x) - E_zg(z)|$ should be no grater than $\sup(\text{MMD}[\mathcal{H}, X, Z])$. Since the squared MMD can be represented by $||\mu_p-\mu_q||_{\mathcal{H}}^2$ and is 0, we can find the second term is 0 for sure. Therefore, for any $\varepsilon>0$, we have 
\begin{equation}
   |E_xf(x) - E_zf(z)| \le 2\varepsilon \text{  for any } f\in C(\mathcal{X}).
\end{equation}
Thus $p=q$ by Lemma 4.2.

\subsection{Proof of Theorem 4.2}
The proof of this theorem relies on the recent advance in ReLU networks universality. Without loss of generalizability, we will assume all the $m$ networks has same maximum width $w$. Given an input domain $\mathcal{X}\subseteq\mathbb{R}^d$ and an output codomain $\mathcal{Y}\subseteq\mathbb{R}$, we define $L^p(\mathcal{X}, \mathcal{Y})$ as the class of $L^p$ functions from $\mathcal{X}$ to $\mathcal{Y}$, endowed with the $L^p$- norm: $||f||_p=(\int_{\mathcal{X}}||f(x)||^p_pdx)^{1/p}$. Park \etal~\cite{park2021minimum} show that, for any $p\in[1, \infty]$, ReLU networks of width $w$ are universal in $L^p(\mathbb{R}^d, \mathbb{R})$ if and only
if $w\ge d+1$. 

We will show that the number of neural network assembled together (i.e., $m$) can be translated into the width of a single neural networks. Thus, we only have to ensure $m\ge\mathrm{ceil}(\frac{d+1}{w})$. 
We assume the inputs as well as outputs  are normalized to $[0, 1]^d$, which can be easily generalized to other cases. 
Following \cite{park2021minimum}, we will construct each neural network into an encoder-memorizer-decoder structure. In the encoder structure, we empoly a quantization function that can quantize a scalar into binary representation. Denote $\mathcal{B}_{n}=\{0, 2^{-n}, 2\times 2^{-n}, \dots, 1-2^{-n}\}$ as the fixed-point set for bit-width $n$, the quantization function is given by:
\begin{equation}
   q_n(x) = Binary(\max\{b\in\mathcal{B}_n:\ b\le x\}),
\end{equation}
where the $Binary$ function means converting the decimal representations into binary representations. For example, scalar 0.5 will be quantized to $100$ using $q_3$ and $1000$ using $q_4$. The quantization function will produce error that is always less than or equal to $2^{-n}$. Thus, the error can be made arbitrarily small by choosing a large $n$. 

For quantization of the input vector $X\in[0, 1]^d$, we will simply concatenate each quantized element in the vector, given by:
\begin{equation}
   \texttt{encode}_n(X) = \sum_{i=1}^d q_n(X_i) \times 2^{-(i-1)n}
\end{equation}
Note that multiplying $2^{-(i-1)n}$ for binary representations is equivalent to perform right shift ${(i-1)n}$ bits. In other words, the inputs vector are encoded into a $dn$-bit binary representation. For the output scalar $y$, we can also encode it into a $k$-bit binary fixed-point number. Now we will construct the memorizer. The memorizer can map each quantized input to each quantized output, given by:
\begin{equation}
   \texttt{memorize}_{n,k}(\texttt{encode}_n(X)) = \texttt{encode}_{k}(y)
\end{equation}
This memorizer function can map one scalar to another scalar. And again, the information loss can be made arbitrarily small by choosing some $n$ and $k$.
Finally, we use a decoder in network to map the binary representation into decimal representation. The decoder can be viewed as the inverse encoder, though we cannot completely restore the original value. 
Han \etal~\cite{hanin2017approximating} prove that any continuous function with $d$-dimensional input to $n$-dimensional outputs can be approximated using ReLU networks of width $d+n$. Therefore, the encoder-memorizer-decoder structure only requires $d+1, 2, 2$ width for each sub-module. Assume a target function $f$, the network function can be constructed as $q_k\circ g \circ q_n$, where $g$ is the memorizer. Thus, for any $\varepsilon>0$, we have
\begin{equation}
   \sup_{x\in[0,1]^d}||f(x)-q_k\circ g \circ q_n(x)||_{\infty}\le \varepsilon,
\end{equation}
by choose large enough $n, k$ so that $w_f(2^{-n})+2^{-k}\le \varepsilon$ where $w_f$ is the modulus of continuity of $f: ||f(x)-f(x^{\prime}||_{\infty}\le w_f(||x-x^{\prime}||_{\infty}) \ \ \forall x,x^{\prime}\in[0,1]^d$.

In fact, the quantization representation, as well as the concatenation of the quantized input, can be assembled from different networks. For example, we can use $km$-bit to encode the output scalar. This implementation gives us flexibility to split the encoded output into $m$ different $k$-bit quantized output scalar. And by simple linear transformation can we merge $m$ outputs into a scalar. The same split-then-merge operation can be applied into the encoder of the input vectors. Each network can concatenate $\frac{d}{m}$ elements in the input vector.

\begin{table*}[h]
   \caption{Pre-trained model zoo for generating MixMix dataset. AD means average down and deep stem layers as proposed in~\cite{he2019bagoftricks}.}
   \centering
   \begin{adjustbox}{max width=0.8\linewidth}
   \begin{tabular}{lclclc}
   \toprule 
   Model & Acc. & Model & Acc. & Model & Acc.\\
   \midrule
   ResNet-34 & 74.03  & MobileNetV2\_1.0 & 73.15 & DenseNet-201 & 77.59 \\
   ResNet-50AD & 79.03 & MobileNetV2\_2.0 & 77.81 & ShuffleNetV2\_1.5 & 72.82\\
   ResNet-101AD & 80.23 & MNasNet\_1.0 & 74.02 & ShuffleNetV2\_2.0 & 74.47\\
   ResNet-152AD & 80.87 & MNasNet\_2.0 & 76.68 & MobileNetV3\_Large\_1.0 & 75.29\\
   RegNetX-600MF & 74.23 & VGG16\_BN & 73.68 & MobileNetV3\_Large\_1.4 & 77.15\\
   RegNetX-1600MF & 77.29 & SE-Net-50 & 79.00 &
   NASNet-7ms & 75.77\\
   RegNetX-3200MF & 78.72 & DenseNet-121 & 75.77 & 
   NASNet-10ms & 77.71 \\
   \bottomrule
   \end{tabular}
\label{tab_modelzoo}
\end{adjustbox}
\end{table*}

As a consequence, each network's memorize will map the split input to the split output. Again, the information loss for each network can be made arbitrarily small by choosing sufficiently large encoding bitwidth. 

\section{Implementation Details}
\subsection{Pre-trained Model Zoo Composition}

We summarize the information in \autoref{tab_modelzoo}.

\subsection{Learning the Loss Weight}

We find the BN statistic loss will vary along with models. This is because the depth and width in each model are not identical. Therefore, it is not practical to manually set the loss weight ($\lambda$ in Eq.~\ref{eq_fmix}) for different models.
To address this problem, we adopt a similar strategy in \cite{kendall2018uncertainty} to learn the loss weight by backpropagation. In particular, we first normalized each loss term to 1 after the first calculation of the loss function. Denote the normalized loss term as $\hat{L}$, the adaptive loss weight is formulated by
\begin{equation}
    \min_{X, \alpha} = \sum_{i} \left(\frac{1}{\alpha_i^2}\hat{L}_i(X) + \alpha_i^2\right), 
    \label{eq_ada_loss_weight}
\end{equation}
where $\hat{L}_i(X)$ is the normalized loss term, e.g. BN statistic loss and cross-entropy loss, and $\alpha_i$ is the learnable loss weight to balance the loss function. Eq.~(\ref{eq_ada_loss_weight}) can tune the loss weight based on the magnitude of each normalized loss term and prevent gradient domination. For example, if $L_i$ becomes two low, $\alpha_i$ will decrease so that the gradient of $\frac{1}{\alpha_i^2} L_i$ will increase accordingly, thus preventing the gradient domination. 
It is worthwhile to note that we do not spend much time in finding the best hyper-parameters and learning rule for $\alpha$, yet the performance of MixMix still outstrips existing methods.

\subsection{Implementation Details}
This section describes the hyper-parameters for image synthesis and data-free compression in detail.

For image synthesis, we use the multi-resolution training pipelines, that is to say, we first downsample the images to $112\times112$ and then use full resolution training to speed-up the training. We use 2.5k iterations for low-resolution optimization and 2.5k iterations for high-resolution optimization. The batch size for each GPU is set to 8. Before forward process of the images, we randomly apply flip and color jitter to the images to simulate the data augmentation. Then we apply Data Mixing ensure exact inversion. The bounding box edge ratio of the data mixing is sampled from $U(0.1, 0.4)$. 
We use Adam optimization with betas to (0.5, 0.5) and do not apply any L2 regularization. For optimization of loss weight, we set a constant learning rate 1e-3. The learning rate of images is set to 0.25 followed by a cosine decay schedule. After each update of the images, we clip the images to prevent the value from increasing too high.

For post-training quantization experiments, we adopt exactly the same hyper-parameters in its original paper~\cite{li2021brecq}. \textsc{Brecq} optimizes the rounding mechanism of weight quantization. It adopts the block-by-block feature reconstruction to optimize the weights, formalized by
\begin{align}
   \argmin_{\mathbf{v}}\ \mathbb{E}[\mathbf{Wx}-\hat{\mathbf{W}}\mathbf{x}] + \lambda \sum_i (1-|2\sigma(\mathbf{v}_i)-1|^{\beta}),\nonumber \\ 
   \mathrm{subject\ to\ \ } \hat{\mathbf{W}} = s \times \mathrm{clip}\left(\lfloor \frac{\mathbf{W}}{s}\rfloor+\sigma(\mathbf{v}), n, p\right) \label{eq_adaround1}. 
\end{align}
The variable $\mathbf{v}$ can determine rounding up or rounding down by using a sigmoid function $\sigma(\cdot)$.
The regularization term in Eq.~(\ref{eq_adaround1}) ensures $\sigma(\mathbf{v})$ can converge to 0 or 1. $n$ and $p$ are the limit integer restricted by bit-width and $\lambda, \beta$ are the hyper-parameters. $\lambda$ is set to 0.1 and $\beta$ will decrease in training process to enhance the regularization.

For quantization-aware training, we use the intermediate feature quantization loss proposed in \cite{haroush2020knowledge} and the loss weight is set to 300. To prevent overfitting of the model, we use random color jitter (0.2, 0.2, 0.2, 0.1), random grad scale (probability=0.2), random Gaussian blur (1, 2, probability=0.5) and random horizontal flip. We freeze the update of BN statistics when the training iteration reaches 10000 since we find updating BN statistics will alter the real activation distribution. The learning rate is set to 0.04 and will be multiplied by 0.1 at iteration 4000, 10000, 20000. Weight decay is set to the same as the full precision model. A recent benchmark paper~\cite{li2021mqbench} gives a comprehensive study of QAT, we do not use such settings since generating 1.2M synthesized images taks too much time. We also avoid studing mixed precision quantization~\cite{cai2020zeroq, li2020efficient} due to the various search space in existing work. We find image quality does not necessarily corresponds to the precision search results. The quality of images should be verified via training.

For data-free pruning experiments, we use a \textit{two-stage} algorithm to prune the network. In stage1, we directly find the weights that has lowest L1 norm, and then add a 0-1 mask to the weights. Then, we jointly train the mask and the weights, given by:
\begin{equation}
   \argmin_{M, \hat{\mathbf{W}}} \mathbb{E}[\mathbf{Wx}-(M\odot\hat{\mathbf{W}})\mathbf{x}]
\end{equation}
After reconstruction, we will absorb the mask and the weights and then find the weights with the lowest L1 norm again. In Stage2, we will only optimize the weights and freeze mask, so that the sparsity will hold after the reconstruction. In each stage we optimize weights or mask for 5k iterations with learning rate 4e-5 followed by a cosine annealing schedule. 

\section{The Choice of Model Zoo}

Sometimes it is not possible to find 20 pretrained models on some dataset. Therefore we evaluate a tiny version of MixMix. 
Here we build two tiny zoo: the first one contains MBV2, MBV3 and MNasNet; the second one contains Res-18, -34, -50. We call them MixMix-mobile and MixMix-residual, respectively. 

Generally, we find add more models can improve the performance compared with single model method. However, we still observed a decrease in generalizability on both MixMix-mobile and MixMix-residual.
Therefore we encorage mixing models from multiple architecture familiy to increase the generalizability.

\begin{adjustbox}{max width=\linewidth}
\begin{tabular}{lccccc}
\hline\hline
Data Source & MobileNetV2 & MNasNet & ResNet-18 & ResNet-50 \\
\hline
Training Set & 64.63 & 58.86  & 69.52 & 74.72 \\
Single-Model & 59.81 & 55.48  & 69.08 & 74.05 \\
MixMix & 64.01 & 57.87 & 70.59 & 74.58 \\
MixMix-mobile & 64.13 & 57.45  & 69.23 & 74.13 \\
MixMix-residual & 61.65 & 56.98 & 69.74 & 74.61\\
\hline
\end{tabular}
\end{adjustbox}

{\small
\bibliographystyle{ieee_fullname}
\bibliography{egbib}
}